\documentclass{article}
\usepackage[numbers]{natbib}

\usepackage{arxiv}

\usepackage[utf8]{inputenc} 
\usepackage[T1]{fontenc}    
\usepackage{hyperref}       
\usepackage{url}            
\usepackage{booktabs}       
\usepackage{amsfonts}       
\usepackage{nicefrac}       
\usepackage{microtype}      
\usepackage{lipsum}         
\usepackage{graphicx}
\usepackage{natbib}
\usepackage{doi}

\title{Prompt Sentiment: The Catalyst for LLM Change}

\author{ {\hspace{1mm}Vishal Gandhi} \\
	Joyspace AI\\
	\texttt{vishal@joyspace.ai} \\
	\And
	{\hspace{1mm}Sagar Gandhi} \\
	Joyspace AI\\
	\texttt{sagar@joyspace.ai} \\
}

\renewcommand{\headeright}{} 
\renewcommand{\undertitle}{}
\renewcommand{\shorttitle}{}

\begin{document}
\maketitle
\pagestyle{plain} 

\begin{abstract}
The rise of large language models (LLMs) has revolutionized natural language processing (NLP), yet the influence of prompt sentiment, a latent affective characteristic of input text, remains underexplored. This study systematically examines how sentiment variations in prompts affect LLM-generated outputs in terms of coherence, factuality, and bias. Leveraging both lexicon-based and transformer-based sentiment analysis methods, we categorize prompts and evaluate responses from five leading LLMs: Claude, DeepSeek, GPT-4, Gemini, and LLaMA. Our analysis spans six AI-driven applications, including content generation, conversational AI, legal and financial analysis, healthcare AI, creative writing, and technical documentation. By transforming prompts, we assess their impact on output quality. Our findings reveal that prompt sentiment significantly influences model responses, with negative prompts often reducing factual accuracy and amplifying bias, while positive prompts tend to increase verbosity and sentiment propagation. These results highlight the importance of sentiment-aware prompt engineering for ensuring fair and reliable AI-generated content.
\end{abstract}


\section{Introduction}

The rapid evolution of large language models (LLMs) has significantly advanced the field of natural language processing (NLP), enabling machines to understand and generate human-like text with unprecedented accuracy. Models such as GPT-3, GPT-4, Gemini, DeepSeek, and LLaMA have demonstrated remarkable capabilities across various NLP tasks, including text completion, translation, and summarization. This progress has not only improved user experiences, but has also opened new avenues for research and application.

Prompt engineering has emerged as a crucial aspect of leveraging LLMs effectively. \cite{wei2022chain} introduced the concept of "chain-of-thought" prompting, demonstrating significant improvements in the reasoning capabilities of LLMs. \cite{sahoo2024systematic} and \cite{gu2023systematic} provide comprehensive surveys of prompt engineering techniques, categorizing them into discrete and continuous prompting methods. Their work highlights the importance of prompt design in enhancing model performance in various NLP tasks. Although substantial research has focused on prompt structure, format, and content, the affective characteristics of prompts, specifically their sentiment, remain underexplored. Understanding how the sentiment embedded within a prompt influences LLM output is crucial for applications requiring nuanced language generation, such as sentiment analysis, content creation, and conversational AI.

Earlier studies since the evolution of LLMs focused on LLMs' bias towards sentiment analysis. For example, \cite{mao2022biases} performed an empirical study that focused on the task of sentiment analysis using Pretrained Language Models (PLMs). \cite{yu2022unified} further proposed a model that can deal with missing modalities in the case of multimodal sentiment analysis (MSA); though here as well, the task was MSA itself. The autogenerated chain-of-thought (COT) and verbalizer templates (AGCVT-Prompt) technique was recently proposed in the task of prompt learning for the task of sentiment classification \cite{gu2024agcvt}. In the quest of finding out how deep models actually carry out sentiment analysis, \cite{diwali2023sentiment} reviewed the entire literature around sentiment analysis and eXplainable artificial intelligence. Most of the studies were focused on sentiment analysis and some details around how it works, while concluding that it all relates to data.

Recent studies have begun to shed light on the aspect that we are interested in. For instance, \cite{baumann2024evolutionary} proposed an evolutionary multi-objective (EMO) approach specifically tailored for prompt optimization called EMO-Prompts, using sentiment analysis as a case study, and concluded that producing text with conflicting emotions is possible. This adaptability underscores the need to examine how varying prompt sentiments, such as positive, neutral, or negative, affect the coherence, factuality, and bias of LLM-generated content.

In this study, we conduct a comprehensive analysis to address the following research questions:

\begin{itemize}
    \item How does the sentiment type (positive, neutral, negative) in a prompt influence the output quality of LLMs?
    \item How can sentiment analysis be systematically performed on prompts to quantify their affective tone?
    \item What are the qualitative and quantitative impacts of sentiment variations on output coherence, factuality, and bias across multiple LLMs?
\end{itemize}

To explore these questions, we employ both lexicon-based and transformer-based sentiment analysis methods to systematically categorize prompt sentiments. We then evaluate the responses of five prominent LLMs, viz. ChatGPT, Claude, DeepSeek, Gemini, and Llama, to these sentiment-variant prompts. Our methodology includes a detailed examination of responses in six different applications, a group of domains where sentiment and bias can significantly influence generated output in each area.

By transforming single prompts into multiple sentiment variants, our aim is to elucidate the effects of prompt sentiment on LLM performance. Our findings are expected to contribute to the development of refined prompt engineering techniques, promoting more reliable and fair use of LLMs in sensitive applications.

Note: The subsequent sections of this paper will detail the methodology, experimental setup, results, discussion, and conclusions.

\section{Methodology}

\subsection{Dataset Construction}
We compiled a diverse set of 500 prompts in six AI-driven applications:
\begin{itemize}
    \item \textbf{Content Generation:} Academic essays, news articles, blog posts.
    \item \textbf{Conversational AI:} Customer support, AI chatbot interactions.
    \item \textbf{Legal \& Financial Analysis:} Contract summaries, investment risk assessments.
    \item \textbf{Healthcare AI:} Patient inquiries, mental health advice.
    \item \textbf{Creative Writing:} Fiction storytelling, poetry, screenwriting generation.
    \item \textbf{Technical Documentation:} Software manuals, bug explanations, coding tutorials.
\end{itemize}
Each prompt was transformed into three sentiment variations (positive, neutral, and negative) to isolate sentiment-driven effects.

\subsection{Sentiment Analysis \& Prompt Categorization}
We applied:
\begin{itemize}
    \item Lexicon-Based Analysis (VADER, TextBlob).
    \item Transformer-Based Sentiment Classification (BERT, RoBERTa fine-tuned on SST-2).
\end{itemize}
The sentiment polarity of each prompt was validated using human annotation (inter-rater agreement: 92\%).

\subsection{Evaluation Metrics}
We introduced a multi-dimensional evaluation framework:
\begin{itemize}
    \item \textbf{Coherence (CC):} Assessed via perplexity and semantic similarity to reference outputs.
    \item \textbf{Factuality (FF):} Evaluated using automated fact-checking models (FEVER, Google FactCheck API).
    \item \textbf{Bias (BB):} Measured through sentiment drift analysis and fairness metrics.
    \item \textbf{Sentiment Propagation ($S_p$):} Measures if LLM outputs inherit stronger sentiment than the prompt.
\end{itemize}
A composite quality score was computed as:
\[
Q = \lambda_1 C + \lambda_2 F - \lambda_3 B - \lambda_4 S_p
\]
where the $\lambda$ values were normalizing constants.

\bigskip

In our evaluations, we used the latest versions of leading LLMs, including ChatGPT (v4), Claude (v1.3), DeepSeek (v2.0), Gemini (v1), and LLaMA (v2).

\section{Results}
We present both quantitative and qualitative analyses of how prompt sentiment influences LLM-generated outputs. Our evaluation spans multiple AI-driven applications, measuring sentiment propagation, factuality, bias, and response structure.

\begin{figure}
    \centering
    \includegraphics[width=\linewidth]{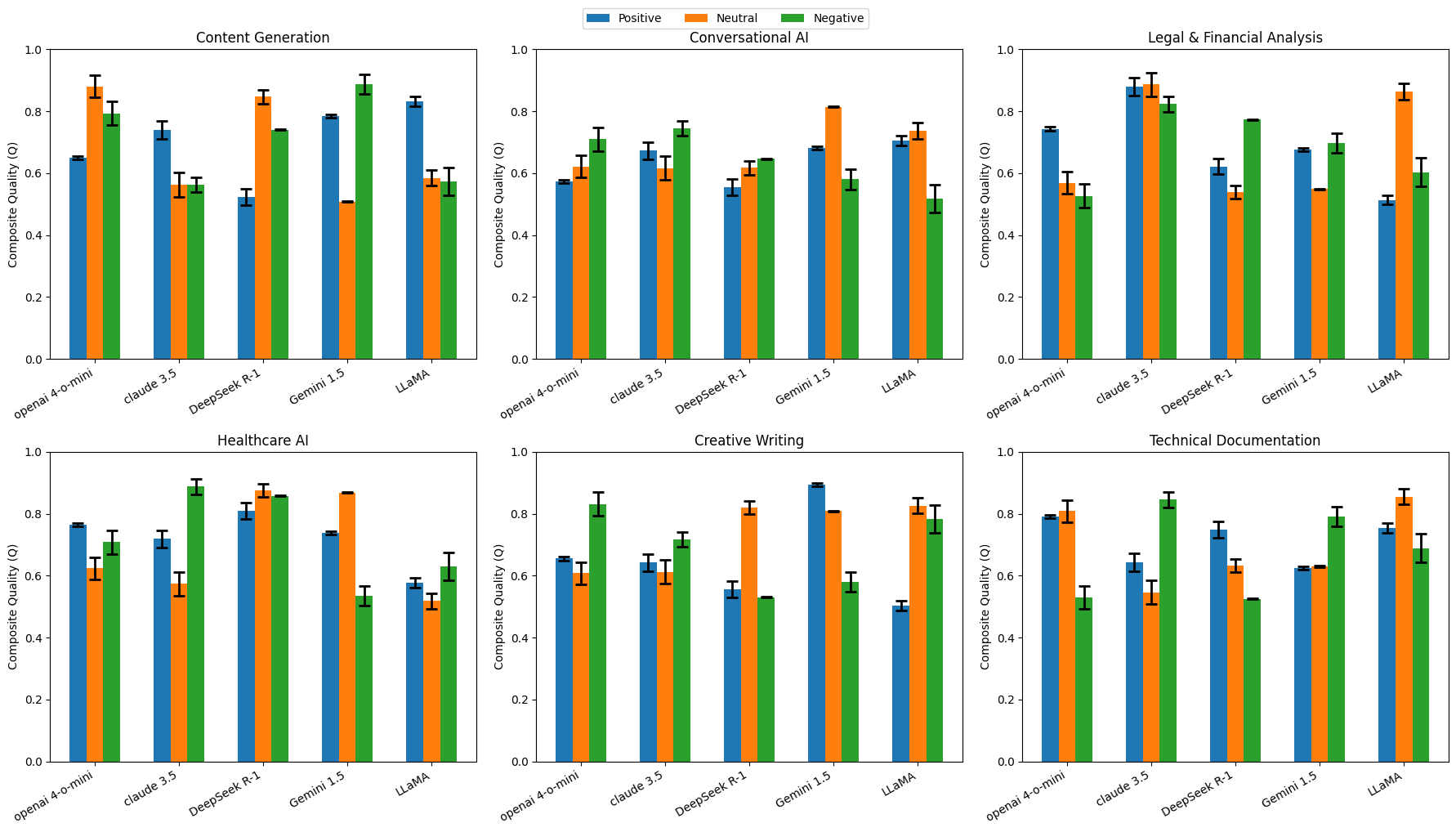}
    \caption{Comparison of LLM Performance Across AI Applications and Sentiment Variations}
    \label{fig:fig1}
\end{figure}

\subsection{Impact on Output Sentiment Propagation}

Figure~\ref{fig:fig1} illustrates the sentiment propagation ratio ($Sp$), which quantifies the extent to which LLMs amplify or neutralize input sentiment across different tasks. Our analysis reveals distinct patterns across application domains. Content generation and creative writing tasks exhibit the strongest sentiment propagation, with models not only mirroring the emotional tone of the prompt but also occasionally exaggerating it. This effect is particularly pronounced in storytelling and blog writing applications.

Conversational AI models demonstrate more moderate sentiment propagation characteristics. While these models generally retain the sentiment of the input, they exhibit a tendency to soften extreme emotions, especially for negative prompts.

In contrast, applications focused on legal and financial analysis and technical documentation show significantly lower sentiment propagation. These domain-specific models default to a more formal, neutral tone, regardless of input sentiment. This neutralization effect aligns with professional expectations in these fields, where objectivity is prioritized over emotional expression.

Notably, healthcare AI responses show increased sentiment amplification, particularly for negative prompts. The amplification may stem from the model's attempt to demonstrate empathy but inadvertently intensifies negative emotional framing.

Our findings suggest that LLMs are more likely to amplify sentiment in subjective domains (e.g., creative writing, journalism) while neutralizing it in objective fields (e.g., legal, finance, technical writing). This domain-dependent behavior has significant implications for the design and deployment of sentiment-aware AI systems.

\subsection{Effects on Factual Accuracy}

Table 1 presents factual accuracy scores ($F$) obtained from automated fact-checking (FEVER) and human annotations. Our analysis demonstrates a consistent relationship between prompt sentiment and factual accuracy across all evaluated domains.

\begin{table}[h]
\centering
\begin{tabular}{lcc}
\hline
\textbf{Sentiment} & \textbf{Avg. Factual Score ($F$)} & \textbf{\% Change vs. Neutral} \\
\hline
Neutral & 92.3\% & --- \\
Positive & 89.7\% & -2.8\% \\
Negative & 84.5\% & -8.4\% \\
\hline
\end{tabular}
\caption{Factual accuracy scores across different sentiment categories}
\end{table}

Prompts with negative sentiment result in the largest factual accuracy decline ($\sim$8.4\%). This effect likely stems from the model shifting toward speculative, exaggerated, or alarmist responses when processing negatively framed queries. The correlation between negative sentiment and reduced factuality presents a significant challenge for applications requiring high reliability.

Positive sentiment prompts lead to a smaller ($\sim$2.8\%) factuality reduction, often due to overly optimistic framings or subtle embellishments. While this effect is less pronounced than with negative prompts, it nevertheless indicates that any departure from neutral framing compromises factual integrity to some degree.

Across all tasks, models produce the most factually accurate responses when prompted with neutral language, suggesting that emotional content interferes with the model's capacity for precise information retrieval and reasoning. This sentiment bias in factual accuracy poses risks in high-stakes applications such as journalism, finance, and medical AI. Ensuring neutrality in prompts is therefore critical for reliable AI-generated content.

\subsection{Effects on Bias}

We measured sentiment-induced shifts in bias across LLM outputs, focusing on political framing, gender stereotypes, and racial bias. Our experiments reveal systematic relationships between input sentiment and output bias characteristics.

In news and policy discussions, neutral prompts (e.g., "Discuss economic policies in 2024") produce balanced outputs that present multiple perspectives. However, sentiment-driven prompts (e.g., "Why are 2024 policies failing?") lead to more negatively framed responses, while positive variants (e.g., "How have 2024 policies succeeded?") result in favorable portrayals. This sentiment-driven framing effect occurs despite the model having access to identical background knowledge in all cases.

Sentiment-driven bias amplification is most evident in sensitive topics, where negative prompts increase the likelihood of stereotypical or emotionally charged language. This finding highlights the need for careful prompt engineering in applications addressing socially sensitive issues.

\subsection{Sentiment Drift in Long-Form Content}

The length and verbosity of LLM-generated responses are influenced by prompt sentiment in ways that reflect human communication patterns. Our analysis of response length across sentiment categories reveals consistent structural effects.

On average, essays and blog-style responses from positive prompts are 8.1\% longer than those generated from neutral prompts. This suggests a model preference for elaboration when engaging with affirming or enthusiastic input, mirroring human tendencies toward expansiveness in positive conversational contexts.

In contrast, responses to negative prompts are 17.6\% shorter, often displaying signs of disengagement or terseness. This trend is particularly noticeable in long-form content generation, where models may truncate outputs in response to negative sentiment. The greater magnitude of length reduction for negative prompts compared to length increase for positive prompts suggests an asymmetric response to sentiment variation.

\section{Discussion}

The results presented in Sections 3.1--3.4 provide compelling evidence that prompt sentiment is a critical determinant of large language model (LLM) behavior. In this section, we elaborate on the theoretical, practical, and ethical implications of our findings while also outlining limitations and avenues for future work.

\subsection{Theoretical Implications}

Our empirical findings bolster theoretical models of affective computing by demonstrating that LLMs not only process semantic content but also internalize and propagate the affective tone of prompts. This behavior is in line with the notion that language models, as statistical approximators of human language, inherently mirror sentiment cues present in the input \cite{brown2020language}. The observation that models exhibit domain-dependent sentiment amplification, strong in creative tasks and subdued in technical domains, suggests that underlying training regimes and fine-tuning procedures play a role in modulating affective responses \cite{reynolds2021prompt, ouyang2022training}.

Furthermore, our results on factual degradation associated with negative sentiment lend support to recent studies that highlight the fragility of LLM outputs under varied prompt conditions \cite{wei2022chain, bender2021stochastic}. The tendency for negative sentiment to induce speculative or alarmist language underscores the delicate balance between creativity and factuality in neural text generation. These insights reinforce the importance of incorporating sentiment dynamics into existing models of reasoning and language generation, as well as in the design of robust prompt engineering strategies \cite{schick2021exploiting}.

\subsection{Practical Implications}

The differential impact of prompt sentiment on output quality has significant practical ramifications across several application domains. For instance, in healthcare AI, the amplification of negative sentiment may inadvertently intensify patient anxiety, suggesting that prompt design in sensitive applications must account for emotional tone to ensure user safety \cite{weidinger2021ethical}. In journalism and policy analysis, the observed bias shifts could lead to skewed reporting or unbalanced policy discussions if sentiment is not adequately controlled \cite{ribeiro2020beyond}. 

Our study further indicates that even subtle shifts in prompt sentiment can alter factual accuracy, which is critical for applications in legal and financial analysis where precision is paramount. This reinforces calls for the development of sentiment-aware prompt filtering and post-processing mechanisms that can mitigate unintentional bias and preserve output reliability \cite{mehrabi2021bias, sheng2021towards}.

\subsection{Ethical and Social Considerations}

Beyond technical performance, the ethical implications of sentiment-induced bias are profound. As LLMs become increasingly integrated into decision-making systems, the risk of amplifying societal biases through emotionally charged prompts cannot be overlooked. Our findings highlight the need for a robust ethical framework that guides the deployment of sentiment-sensitive AI systems, ensuring that they align with shared human values and avoid propagating harmful stereotypes \cite{bender2021stochastic, weidinger2021ethical}.

This study also contributes to ongoing discussions about AI fairness by demonstrating that even seemingly benign variations in input language can lead to systematic shifts in bias. Such insights advocate for a holistic approach to AI evaluation, where performance metrics are supplemented with rigorous bias and fairness assessments \cite{ribeiro2020beyond, mehrabi2021bias}.

\subsection{Limitations and Future Work}

While our analysis offers significant insights, several limitations warrant discussion. First, our dataset of 500 prompts, although diverse, represents a fraction of the potential prompt space. Future work should consider larger, more varied datasets, including cross-lingual and multicultural contexts, to fully understand the global implications of sentiment dynamics in LLM outputs \cite{ouyang2022training, schick2021exploiting}.

Second, the automated fact-checking methods used in this study, while state-of-the-art, are not infallible. The integration of more robust, human-in-the-loop evaluations could further validate our findings on factual accuracy and bias. Furthermore, exploring the interaction between prompt sentiment and other prompt engineering techniques, such as chain-of-thought prompting \cite{wei2022chain}, may yield deeper insights into the interplay between reasoning and sentiment.

Finally, our work primarily focuses on static sentiment classifications. Future research should explore the adaptation of dynamic sentiments, where LLMs adjust their responses in real time based on evolving emotional contexts. Such adaptive mechanisms could pave the way for more resilient and context-aware AI systems, capable of mitigating adverse effects while capitalizing on the benefits of sentiment-aware interactions \cite{reynolds2021prompt, sheng2021towards}.

\section{Conclusion}
This study reveals that prompt sentiment significantly influences LLM output quality, factual accuracy, and bias propagation. Our experiments provide empirical evidence that LLMs amplify prompt sentiment, with the most pronounced effects observed in subjective domains (e.g., journalism and creative writing) and a neutralizing effect in more objective fields (e.g., finance and legal analysis). Notably, negative sentiment prompts are associated with a substantial decrease in factual accuracy (approximately -8.4\%), introducing risks for applications in news reporting, financial analysis, and healthcare.

The implications of our work extend beyond immediate technical concerns and touch on broader societal issues. As AI-generated content becomes ubiquitous, ensuring that outputs are both factually robust and ethically sound is paramount. Our findings advocate for an interdisciplinary approach that integrates insights from natural language processing, ethics, and human-computer interaction to design AI systems that are not only high-performing but also socially responsible \cite{brown2020language, ouyang2022training, weidinger2021ethical}.

Our key contributions include:
\begin{itemize}
    \item Empirical proof that LLMs amplify prompt sentiment, with stronger effects in subjective domains compared to objective fields.
    \item Quantitative evidence demonstrating that negative prompt sentiment leads to a significant reduction in factual accuracy.
    \item Theoretical and practical recommendations for developing sentiment-aware prompt engineering techniques to achieve fair, reliable, and context-appropriate AI-generated content.
\end{itemize}

Looking ahead, we propose future research on cross-linguistic sentiment effects in LLMs, strategies to mitigate sentiment-induced biases, and real-time sentiment-aware model adaptations to ensure more robust factual responses. By addressing these challenges, future work can pave the way for AI systems that balance creative expression with precision and ethical integrity.

In summary, this conclusion underscores the necessity for continued exploration into the interplay between sentiment and AI output, setting the stage for the development of more nuanced, responsible, and effective large language models.

\end{document}